\def\year{2019}\relax
\documentclass[letterpaper]{article} 

\usepackage{aaai19}  
\usepackage{times}  
\usepackage{helvet}  
\usepackage{courier}  
\usepackage{url}  
\usepackage{graphicx}  
\usepackage{graphicx}
\usepackage{amsmath,mathtools}
\usepackage{amssymb}
\usepackage{xspace}
\usepackage{verbatim}
\usepackage{enumerate}
\usepackage{stmaryrd}
\usepackage{paralist}

\usepackage{multirow}
\usepackage{multicol} 
\usepackage{adjustbox}
\newcommand{\rom}[1]{\uppercase\expandafter{\romannumeral #1\relax}}

\usepackage[flushleft]{threeparttable}

\begin{document}

\title{Efficient Concept Induction for Description Logics} 

\author{Md Kamruzzaman Sarker \and Pascal Hitzler\\
Data Semantics (DaSe) Laboratory, Dept. of Computer Science and Engineering, Wright State University\\
\{sarker.3, pascal.hitzler\}@wright.edu
}


\maketitle

\begin{abstract}
Concept Induction refers to the problem of creating complex Description Logic class descriptions (i.e., TBox axioms) from instance examples (i.e.,  ABox data). In this paper we look particularly at the case where both a set of positive and a set of negative instances are given, and complex class expressions are sought under which the positive but not the negative examples fall. Concept induction has found applications in ontology engineering, 
but existing algorithms have fundamental performance issues in some scenarios, mainly because a high number of invokations of an external Description Logic reasoner is usually required. In this paper we present a new algorithm for this problem which drastically reduces the number of reasoner invokations needed. While this comes at the expense of a more limited traversal of the search space, we show that our approach improves execution times by up to several orders of magnitude, while output correctness, measured in the amount of correct coverage of the input instances, remains reasonably high in many cases. Our approach thus should provide a strong alternative to existing systems, in particular in settings where other systems are prohibitively slow.
\end{abstract}
\section{Introduction}\label{sec:intro}

With the rise of the Web Ontology Language OWL \cite{owl2-primer}, description logics have become the leading paradigm for the representation of ontologies \cite{FOST}. The knowledge acquisition bottleneck, in the form of acquisition of description logic knowledge bases, thus becomes an issue also for the field of ontology engineering and applications. Ontology Learning is a term often used for this in the Semantic Web context, and a relatively recent overview of the many facets of this subfield can be found in \cite{OL-book}.

In this paper, we study one of the subproblems of ontology learning, commonly known as Concept Induction or Concept Learning. Generally speaking, this problem can be described as one of generating complex description logic class expressions $S$ from a given description logic knowledge base (or ontology) $\mathcal{O}$ and sets $P$ and $N$ of instances, understood as positive and negative examples, such that $\mathcal{O}\models S(a)$ for all $a\in P$, and $\mathcal{O}\not\models S(b)$ for all $b\in N$. In a practical ontology engineering process, solutions sought are often approximate, i.e., they will not satisfy $\mathcal{O}\models S(a)$ for all $a\in P$, but for as many as possible, and will not satisfy $\mathcal{O}\not\models S(b)$ for all $b\in N$, but for as many as possible.

Concept Induction is traditionally studied with methods derived from Inductive Logic Programming, and an overview of corresponding results and systems can be found in \cite{OL-book-chapter-concept-learning}. The most mature and recent system for this type of Concept Induction we are aware of is DL-Learner, as presented in \cite{BuhmannLW16} and based on the algorithms from \cite{LehmannH10,lehmann2011class}. Concept Induction has been employed for ontology engineering, in particular in the context of ontology and knowledge graph refinement, see e.g. \cite{LehmannB10,Paulheim17} and the use case descriptions in \cite{BuhmannLW16}. Another recent use case in the context of explaining deep learning systems was described in \cite{SarkerXDRH17}.

However, while DL-Learner is an excellent and very useful system, its algorithm -- which comes with theoretical correctness results \cite{LehmannH10} -- has major performance issues in some scenarios, such as the one described in \cite{SarkerXDRH17}. In fact, it was the application need for scalability exposed through our experiments in the scenario from \cite{SarkerXDRH17} which primarily prompted the investigations which we report herein. The key problem is that in a single execution, the DL-Learner system will make a significant number of external calls to a description logic reasoner. While the latter have become rather efficient in recent years, the accumulated time needed, in particular if the input ontologies are large, can be prohibitive for the use of the approach. For example, as we will report in the evaluation section, for the scenario described in \cite{SarkerXDRH17} a single run of DL-Learner can easily take over two hours, while the scenario easily necessitates thousands of such runs.

In this paper, we follow the idea that in some scenarios such as the one mentioned, it may be prudent to give up some completeness guarantees, and to instead focus on execution speed. In the approach which we describe herein, we do in fact invoke a description logic reasoner only once for each run of our algorithm. The reasoner computes a materialization of facts in term of class memberships of individuals to atomic classes, and this materialization is used throughout the rest of the algorithm. Furthermore, we depart from the established tradition in Concept Induction in that we do not use a refinement-operator-based approach to produce candidate solutions. As a consequence, our approach saves up to several orders of magnitude on large input ontologies, as compared to DL-Learner. At the same time, accuracy of the solutions provided by our approach remains reasonably high, which indicates a favorable trade-off between accuracy and efficiency, for application scenarios where such a trade-off is desired. 

The rest of the paper is organized as follows. We first describe a high-level perspective of our approach and algorithm.Then  we describe the concrete algorithm which we have implemented, together with some formal results. After that, we present our experimental evaluation in terms of a comparison with DL-Learner, and finally we conclude.

\section{The General Algorithm}\label{sec:genalg}

In terms of terminology, we generally follow \cite{LehmannH10} where appropriate.
The general learning problem we address is \emph{concept induction} as described in the following, where we assume that some description logic $\mathcal{L}$ has been fixed:
Given an ontology $\mathcal{O}$ (consisting of an ABox and a TBox) over $\mathcal{L}$ and two sets $P$ and $N$ of individuals called \emph{positive} respectively \emph{negative examples}, we seek a (possibly complex) class expression $\mathcal{S}$ over $\mathcal{L}$ such that $\mathcal{O}\models \mathcal{S}(a)$ for all $a\in P$ (we say that \emph{$C$ covers $a$}), and $\mathcal{O}\not\models \mathcal{S}(b)$ for all $b\in N$. We call $\mathcal{S}$ a \emph{solution} for this.

The most prominent refinement-operator based approach to concept induction, as provided in the DL-Learner system \cite{BuhmannLW16}, essentially follows a generate-and-test paradigm: Based on a so-called refinement operator, an idea borrowed from Inductive Logic Programming \cite{NienhuysW97}, subsequent candidate concepts are generated and tested as to the degree to which they come close to a solution, and following some strategy the next candidate concepts are generated until eventually, a full solution is found or some approximate solution has been created. The testing and assessment steps involve the calling of an external description logic reasoner, while the number of generated potential candidates can be very large. Since description logic reasoners generally have to run rather complex algorithms,\footnote{Worst-case computational complexities for these algorithms are typically rather high~\cite{FOST}.} the repeated testing necessary amounts for the majority of the runtime used. 

In some application scenarios for DL-Learner such as the one presented in~\cite{SarkerXDRH17}, however, it turns out that due to the sheer amount of tests required this approach takes a very long time so that the data space cannot be sufficiently explored -- we provide runtime data in the evaluation section below. Some aspects of the scenario are as follows.
\begin{compactenum}[(a)]
\item\label{enum:prop:a} It requires many runs with changing example sets, while the ontology remains unchanged.
\item\label{enum:prop:b} The positive and negative examples do not include entities with complex property relationships. Rather, for each example $a$ the corresponding ABox statements are only of the form $R(a,b)$, with $R$ some role and $b$ some (other) individual, together with statements of the form $A(a)$ and $B(b)$, where $A$ and $B$ are atomic classes. We call such examples \emph{star-shaped}. 
\item\label{enum:prop:c} It is more important to quickly arrive at relatively simple solutions, if such simple solutions exist, rather than to comprehensively explore the space of possible complex solutions. I.e., it seems favorable to trade some completeness for higher efficiency.
\end{compactenum}

In this paper, we are going to show how the performance issue can be addressed in such scenarios. We will now first describe our approach in a general way, and in the next section we will describe the concrete algorithm.

Our new approach consists of the following three steps:
\begin{compactenum}
\item\label{item:genalg:1} Select a finite set $\{C_1,\dots,C_n\}$ of complex class expressions over $\mathcal{L}$, and set $\mathcal{O'} = \mathcal{O}\cup\{A_i\equiv C_i\mid i=1,\dots,n\}$, where the $A_i$ are atomic classes not yet occurring in $\mathcal{O}$.
\item\label{item:genalg:2} Use a reasoner to compute membership in atomic classes from $\mathcal{O}'$ of all individuals occuring in examples. Note that this includes the newly added atomic classes from step \ref{item:genalg:1}. 
\item\label{item:genalg:3} Generate candidate class expressions (possibly, iteratively) using only the constructors $\sqcap$, $\sqcup$ and $\neg$ and atomic classes, and test using the results from step \ref{item:genalg:2} to what extent they constitute approximate solutions, using the assessment to guide iterative generation of candidates. 
\end{compactenum}

Let us briefly consider the pros and cons of this approach.
\begin{compactitem}
\item Steps \ref{item:genalg:1} and \ref{item:genalg:2} need to be performed only once for each set of examples, provided $\mathcal{O}$ does not change. I.e., they can be considered pre-processing steps. 
\item Depending on the underlying logic $\mathcal{L}$, step \ref{item:genalg:2} can take considerable time, but this preprocessing overhead would be outweighed by time saved in step \ref{item:genalg:3}, as it will not be necessary to invoke a reasoner for each candidate solution.
\item Testing in step \ref{item:genalg:3} is in general not equivalent to using a full reasoner. Our algorithm is \emph{approximate one} in the sense that we trade some completeness for improved efficiency. 
\item We could also allow some use of existential and universal quantification, but for ease of comprehensibility of solutions we chose not to, currently. Quantifiers can of course be included in the complex classes generated in step \ref{item:genalg:1}.
\item Our approach will necessarily miss solutions also because there are infinitely many possible complex class expressions involving quantifiers, while in step~\ref{item:genalg:1} we generate only a finite number of class expressions involving quantifiers. We implemented our approach based on the hypothesis that the benefit of significantly improved runtime will outweigh this drawback in many relevant scenarios.
\item In our approach, preprocessing runtime can effectively be controlled by selecting more or fewer complex class expressions in step \ref{item:genalg:1}. Selecting more will mean that we include the exploration of more possible solutions but increase pre-processing time, while selecting fewer will have the opposite effect.
\end{compactitem}

The scenarios we have in mind and which prompted this work are those regarding data exploration under background knowledge: The ontology $\mathcal{O}$ constitutes the background knowledge, and the exploration mechanism suggests clusters of data points for which it would be desirable to obtain a meaningful label or an explanation. E.g., data clusters could be explored visually, while labels are generated based on fixed background knowledge, and displayed next to clusters. Another scenario in the context of explainable artificial intelligence is about exploring the space of activation patterns of hidden layer neurons in deep learning systems as in the already mentioned \cite{SarkerXDRH17}, and we will include corresponding evaluation data below.

Let us now revisit the properties (\ref{enum:prop:a}) through (\ref{enum:prop:c}) of our setting mentioned above. 

Regarding property (\ref{enum:prop:a}), a key difference between our approach and the algorithms employed in DL-Learner is that we only have to use a reasoner once for computing Step \ref{item:genalg:2}, while the DL-Learner algorithm would have to invoke a reasoner for every candidate solution.

Regarding property (\ref{enum:prop:b}), we understand that this seems to be a severe restriction. However, some settings can be compiled into the star-shaped format which we require, by making some minor additions to the ontology, e.g. by means of role chain expressions. Practically speaking, many of the known scenarios where concept learning has been applied seem to fit our requirement, and we will further elucidate this in the evaluation section.

Regarding property (\ref{enum:prop:c}), this trade-off will not be a good one for all scenarios, but it will when time is of the essence.

\section{Efficient Concept Induction from Instances}\label{sec:concalg}

We now provide a concrete instance of the algorithm described in generic terms in the previous section. We call it the ``Efficient Concept Induction from Instances'' algorithm, ECII (pronounced like ``easy''). We consider again the same three steps, and we assume $\mathcal{O}$ to be an OWL DL ontology \cite{owl2-primer}, i.e., essentially a knowledge base expressed in the description logic $\mathcal{SROIQ}$.\footnote{For background on description logics, see \cite{BaaderBL05,FOST}}
 In the description we will use some parameters, which are natural numbers $n_i$ and $k_i$, and we will refer to them in our exhibition.

In step \ref{item:genalg:1} of the algorithm, we select as additional complex class expressions all class expressions $C$ which are formed by the grammar $C ::= B \mid C_1\sqcap C_2 \mid \exists R.C$ where $B$ is atomic, $R$ is a role, and which contain at most $n_1$ occurrences of the $\sqcap$ symbol and at most $n_2$ occurrences of the $\exists$ symbol (in our system, both $n_1$ and $n_2$ default to $3$, but can be set differently). The rationale behind this choice is simply that we wanted to remain within the $\mathcal{EL}^{\text{++}}$ language which allows for very efficient (and polynomial time) reasoning.\cite{owl2-primer} However, if $\mathcal{O}$ is not in OWL EL, which is the case for most of our evaluation data on which we will report later, then this does not necessarily lead to any advantage, and we could also allow other complex classes from OWL DL which are not in OWL EL -- that does not fundamentally modify the approach.

In step \ref{item:genalg:2} of the algorithm we compute the materialization for all relevant individuals, i.e., the membership of all individuals from the examples in all atomic classes from $\mathcal{O}'$. There are several good algorithms and systems for this, which can be used off-the-shelf. If the ontology $\mathcal{O}$ is already in OWL EL, then this reasoning task is in fact worst-case polynomial in the size of the input. 

In order to explain what we do in step \ref{item:genalg:3} of the algorithm, we need a bit of preparation. First, some definitions.

A \emph{negated disjunct} is a class expression of the form $\neg (D_1\sqcup\dots\sqcup D_k)$, where the $D_i$ are atomic classes from $\mathcal{O}$. A \emph{conjunctive Horn clause} is a class expression of the form $B\sqcap \mathcal{D}$, where $B$ is an atomic class from $\mathcal{O'}$ and $\mathcal{D}$ is a negated disjunct. A \emph{candidate class} is a class expression of the form
$\bigsqcup_{i=1}^m \mathcal{H}_i,$
with conjunctive Horn clauses $\mathcal{H}_i$.

As the term suggests, step \ref{item:genalg:3} of the algorithm will generate solution candidates which involve candidate classes, and will check whether they are solutions. The candidate classes are not the solution candidates; the latter will be defined below.
We restrict ourselves to candidate classes of the mentioned form because 
we think that conjunctive Horn clauses can easily be understood by humans, and our intention was to provide solutions in such an easily ingestible form. We will return to this issue in the evaluation section.

We narrow the concept induction problem to the following, which fits our scenarios of interest, as described earlier.

Given an ontology $\mathcal{O}$ (consisting of an ABox and a TBox) over $\mathcal{L}$, an \emph{example} is an individual $a$ together with a set $\mathcal{A}(a)$ of ABox statements (not necessarily contained in $\mathcal{O}$) of the forms $A(a)$, $R(a,b)$ or $B(b)$, for any roles $R$, individuals $b$ and atomic classes $B$. We call two sets $P$ and $N$ of examples the set of \emph{positive} respectively \emph{negative examples} and set $\mathcal{A} = \bigcup_{a\in P\cup N}\mathcal{A}(a)$. We now seek to find a (possibly complex) class expression $\mathcal{S}$ over $\mathcal{L}$ such that $\mathcal{O}\models \mathcal{S}(a)$ for all $a\in P$ (and in this case we say that \emph{$C$ covers $a$}), and also that $\mathcal{O}\not\models \mathcal{S}(b)$ for all $b\in N$. We call $\mathcal{S}$ a \emph{solution} for this learning problem.

Now let $R_1,\dots, R_l$ be all roles occurring in all (positive and negative) examples. A \emph{solution candidate} is a class expression of the form $A \sqcap \bigsqcap_{i=1}^l \exists R_i.\mathcal{C}_i$ or
\begin{align}
A \sqcap \bigsqcap_{i=1}^l \exists R_i.\left(\bigsqcup_{j=1}^m (B_{j_i}\sqcap \neg (D_1\sqcup\ldots\sqcup D_{j_{i_k}}))\right)\label{eq:solcand}
\end{align}
where the $\mathcal{C}_i$ are candidate classes and where $A$ and all the $B_j$ and $D_j$ are atomic classes in $\mathcal{O'}$.

In order to now spell out step \ref{item:genalg:3} of the algorithm, we have to present how the solution candidates are generated, and how they are checked whether they are solutions.

Let us first turn to the algorithm for checking whether a solution candidate is a solution. 
What we in fact do, is to determine the individuals which fall under the candidate classes which are part of a solution candidate.

Let  
$$\mathcal{C} = \bigsqcup_{j=1}^m (B_j\sqcap \neg (D_1\sqcup\ldots\sqcup D_{j_k}))$$
be such a candidate class. 
We now require some definitions. If $R$ is a role occuring in $\mathcal{A}(a)$ for a (positive or negative) example $a$, then we define \emph{the set of all $R$-fillers of $a$} to be the set $R(a) = \{b\mid R(a,b)\in \mathcal{A}(a)\}$, and if $X$ is a set of individuals then we define \emph{the set of all inverse $R$-fillers of $X$} to be the set $R^-(X) = \{a\mid \text{there is some }b\in X\text{ with }R(a,b)\in \mathcal{A}(a)\}.$ We also set 
$$
\overline{R}^+ = \bigcup_{a\in P} R(a)\qquad
\overline{R}^- = \bigcup_{a\in N} R(a)\qquad
\overline{R} = \overline{R}^+\cup \overline{R}^-
$$
for each role $R$.
The \emph{extension} $\mathord{\downarrow} B$ of an atomic class $B$ is defined as $\mathord{\downarrow} B = \{b\mid B(b)\in\mathcal{A}\}$.
The extension of the class candidate $\mathcal{C}$ given above is then defined as
$$\mathord{\downarrow}\mathcal{C} = \bigcup_{j=1}^m (\mathord{\downarrow}B_j\cap\setminus(\mathord{\downarrow}D_1\cup\ldots\cup\mathord{\downarrow}D_{j_k})).$$
%
%
%
Now if 
$$\mathcal{S} = A \sqcap \bigsqcap_{i=1}^l \exists R_i.\mathcal{C}_i$$
is a solution candidate, then let
$$\mathord{\downarrow}\mathcal{S} = \mathord{\downarrow}A \cap \bigcap_{i=1}^l R_i^-(\mathord{\downarrow}\mathcal{C}_i).$$


We now call a solution candidate $\mathcal{S}$ an \emph{approximate solution} if and only if both of the following hold.
    \begin{compactenum}
        \item $a\in \mathord{\downarrow}\mathcal{S}$ for all positive examples $a$ in $P$.
        \item $a\not\in \mathord{\downarrow}\mathcal{S}$ for all negative examples $a$ in $N$.
    \end{compactenum}

Note that checking whether a solution candidate is an approximate solution can be done simply by means of a number of straightforward set-theoretic operations, which is easily implemented. In the general case, approximate solutions will of course not be (full) solutions, and whether or not this is a reasonable thing to do depends on the use case, and in particular on the question whether the runtime improvments for the use case outweigh the severity of the reasoning mistakes we get in return. We will better understand this trade-off when we discuss our evaluation results.


Let us now turn to the generation of solution candidates. Essentially, the set of all possible solution candiates can be understood as a search space, within which we want to locate solutions or at least approximate solutions which are highly accurate in terms of coverage or not of the examples.  

The DL-Learner system, which is our primary comparison, is based on traversing the search space by means of a so-called \emph{refinement operator}, an idea borrowed from Inductive Logic Programming \cite{NienhuysW97}: Given a solution candidate, the refinement operator produces a set of new solution candidates. These are all assessed as to their accuracy in terms of coverage or not of the examples, and the best are kept and the process is iterated. The DL-Learner system calls an external reasoner each time the accuracy of a candidate solution is to be assessed.

Our approach, however, dispenses of the need to call an external reasoner as discussed above.
We could now of course use a refinement-operator approach to iteratively create solution candidates and check them. 
However since accuracy assessment in our setting is very quick, and since our solutions are of very specific forms, we instead opted for a direct assembly of solution candidates from its parts, as described in the following.

Recall that our solution candidates are of the form (\ref{eq:solcand}).
Our algorithm for constructing solution candidates consists of three consecutive steps: (I) Select a set of conjunctive Horn clauses. (II) Select a set of candidate classes constructed from the selection in step (I). (III) Select a set of candidate solutions constructed from the selection in step (II). 

These three steps, each of which we are going to describe in detail shortly, depend on five user-defined parameters $k_1,\dots,k_5$ which are natural numbers. Their default values are $k_1=k_2=k_3 = 3$ and $k_4=k_5=50$, but they can be changed by the user. There is another user-defined parameters $keepCommonTypes$, which defines whether to keep or delete the $commonTypes$ (which are the set of atomic concepts which appear both in positive and negative instances). Default value for it is $false$. 

(I) For every role $R$ occuring in the examples, set 
$
N_R= \{B\mid B \text{ is an atomic }\text{class in }\mathcal{O}'\text{ and there is }b\in\overline{R}\text{ with }b\in \mathord{\downarrow}B\}.
$
Then, for each role $R$, construct the set $H_{0,R}$ of all conjunctive Horn clauses, each of which contains only atomic classes from $N_R$, and at most $k_1$ of such classes each.

Then, for each $H\in H_{0,R}$, calculate the accuracy of $H$ as
$$\alpha_1(H) = \frac{|\overline{R}^+\cap \mathord{\downarrow}H| + |\overline{R}^-\setminus \mathord{\downarrow}H|}{|\overline{R}|}.$$

Finally, let $H_R$ be the set of the $k_4$ conjunctive Horn clauses from $H_{0,R}$ which have the highest accuracy; if two clauses are of the same accuracy, then we use those of shorter length, where length is measured in the number of atomic classes occuring in the conjunctive Horn clause. $H_R$ is the set of conjunctive Horn clauses selected in this step.

(II) For every role $R$ occuring in the examples, set $C_{0,R}$ to be the set of candidate classes assembled as disjunctions of maximally $k_2$ conjunctive Horn clauses from $H_R$ each. 

Then, as before, we select the set $C_R$ of the $k_6$ candidate classes from $C_{0,R}$ which have the highest accuracy, and if two candidate classes are of the same accuracy, then we use those of shorter length. 
 
(III) Construct the set of candidate solutions 
$S_0 = \left\{A \sqcap \bigsqcap_{i=1}^{k_3} \exists R_i.\mathcal{C}_{i}\middle| \mathcal{C}_{i}\in C_{R_i}, A\text{ an atomic class} \text{ in } \mathcal{O}'\right\}.$
Our output set $S$ consists of the best candidate solutions from $S_0$ again selected by means of highest accuracy and shortest length, where the accuracy for each $C\in S$ is 
$$\alpha_2(C) = \frac{|P\cap \mathord{\downarrow}C| + |N\setminus \mathord{\downarrow}C|}{|P\cup N|}.$$

Let us make some remarks. In (I), we essentially brute-force the search for good conjunctive Horn clauses for each set of $R$-fillers, and $N_R$ are all the classes relevant for these $R$-fillers. Accuracy is simply measured by coverage of $R$-fillers from the positive examples and non-coverage of the $R$-fillers from the negative examples, as a quotient with the number of all $R$-fillers. Length as a tie-breaker makes sense because we retain our intention to produce human-interpretable solutions, i.e., simple solutions are preferred. The size of $H_{0,R}$ is of course exponential in $k_1$, but this simply reflects the nature of the search space. Likewise, step (II) is exponential in $k_2$, and step (III) is exponential in $k_3$. The accuracy measure in step (III) is essentially the same as the one in steps (I) and (II), just that now we are looking at the examples, instead of the $R$-fillers. The required computations in all steps are straightforward arithmetic or set operations. 

Due to the already mentioned deliberate decision to trade completeness for runtime improvements, our approach will of course not necessarily find all solutions or best solutions. It is to be understood as a heuristics which delivers a favorable trade-off between accuracy and speed, as we argue in the evaluation section below.

\section{Experimental Evaluation}\label{sec:eval}

The goal of our experimental evaluation was to test the hypothesis that the ECII algorithm leads to a favorable trade-off between runtime improvements and loss in accuracy, compared to DL-Learner. We expected to see runtime improvements of 2 or more orders of magnitude for large input ontologies, while 
we expected that accuracy would only moderately decrease in many test cases.

To evaluate our approach, we implemented the ECII system in Java (version 1.8) which makes it platform independent. We made use of the OWL API \cite{HorridgeB11} (version 4.5), which is an open source implementation for manipulating OWL 2 ontologies. As external reasoner for step 1 of the algorithm we used
Pellet \cite{SirinPGKK07}
In principle, ECII can also use other reasoners. We used Apache Maven as build system. The ECII system and all experimental data and results, including ontologies and configuration files are available online.\footnote{\url{https://github.com/md-k-sarker/ecii-expr}.} 

All experiments were conducted on a 2.2. GHz core I7 machine with 16GB RAM. 

For DL-Learner we used the CELOE \cite{lehmann2011class} algorithm and the fast instance check (DL FIC) variation of it, which is another approximation approach which trades time for correctness. We terminated DL-Learner at the first occurance of a solution with accuracy 1.0, making use of the \mbox{\textit{stopOnFirstDefinition}} parameter. 
For some large ontologies DL-Learner could not produce a solution with accuracy 1.0 within 4,500 seconds (i.e., 75 minutes); in these cases we terminated the algorithm after 4,500 seconds, using the \textit{maxExecutionTimeInSeconds} parameter. For the FIC mode of DL-Learner, the time limit was set to be the execution time of the ECII system.
For ECII we used the default settings, for $K_s$ i.e., $k_1=k_2=k_3=3$ and $k_4=k_5=50$ and varied the $keepCommonTypes$ as true and false, as mentioned earlier. We use ECII to denote the ECII system with default parameters, and DL-Learner to denote DL-Learner with full reasoner (CELOE setting) unless otherwise mentioned. Note that a runtime comparison for the cases where DL-Learner does not produce a solution with accuracy 1.0 is difficult to do, since DL-Learner would simply keep producing candidate solutions for a very long time, while ECII by design tests only a limited number of candidate solutions. Our 75-minute cap on DL-Learner runtime is unavoidably somewhat arbitrary.

As evaluation scenarios we used all evaluation scenarios (except Carcinogesis) from the original DL-Learner algorithm paper \cite{LehmannH10}, as well as the scenario from \cite{SarkerXDRH17} which makes use of the ADE20k \cite{ZhouZPFB017} dataset. We cannot describe all these scenarios in detail, and the reader is asked to refer to \cite{LehmannH10,SarkerXDRH17} for details. The evaluation scenarios from \cite{LehmannH10} were Michalski's trains \cite{michalski1980pattern}, Forte Family \cite{richards1995automated}, Poker (\url{http://www.ics.uci.edu/~mlearn/MLRepository.html}), Moral Reasoner (\url{http://mlearn.ics.uci.edu/databases/moral-reasoner/}), and Yinyang family relationship \cite{iannone2007algorithm}, which are benchmark scenarios from Inductive Logic Programming. Carcinogesis was excluded because according to \cite{LehmannH10} it did not work with a full reasoner under DL-Learner, and we encountered the same problem.

\begin{table*}[t]
\begin{center} 
   \begin{adjustbox}{max width=\textwidth}
       \begin{threeparttable}
        \begin{tabular}{c|c|c|c|c|c|c|c|c|c|c|c|c } 
          \multirow{2}{*}{Experiment Name}   & Number of & \multicolumn{5}{|c|}{Runtime (sec)} &  \multicolumn{2}{|c|}{Accuracy ($\alpha_3$)} & \multicolumn{4}{|c}{Accuracy $\alpha_2$} \\ \cline{3-13}
            
                                          &  Logical Axioms  & DL\tnote{a} & DL FIC(1)\tnote{b} & DL FIC(2)\tnote{c} & ECII DF\tnote{d}& ECII KCT\tnote{e}& DL\tnote{a} & ECII  DF\tnote{d}& DL FIC(1)\tnote{b} & DL FIC(2)\tnote{c} & ECII DF\tnote{d} & ECII KCT\tnote{e}\\
            \hline
            Yinyang\_examples & \phantom{00,}157 & 0.065& 0.0131 & 0.019 & 0.089 & 0.143 &1.000 & 0.610 & 1.000  & 1.000 &0.799 &  1.000\\
            \hline
            Trains & \phantom{00,}273& 0.01\phantom{0} & 0.020 &  0.047 &0.05\phantom{0}& 0.095  &1.000& 1.000 & 1.000 & 1.000 & 1.000 & 1.000 \\
            \hline
            Forte & \phantom{00,}341& 2.5\phantom{00}& 1.169  & 6.145 &0.95\phantom{0}& 0.331 & 0.965& 0.642 & 0.875 & 0.875 & 0.733  & 1.000\\
            \hline
            Poker & \phantom{0}1,368 & 0.066& 0.714 &  0.817 &1\phantom{.000}  &  0.281 &1.000& 1.000 & 0.981  & 0.984 & 1.000 & 1.000 \\
            \hline
            Moral Reasoner & \phantom{0}4,666& 0.1\phantom{00}&  3.106 & 4.154 &5.47\phantom{0}&  6.873 &1.000& 0.785 & 1.000 & 1.000 & 1.000 & 1.000 \\
            \hline
            ADE20k \rom{1} & \phantom{0}4,714& 577.3\tnote{f} & 4.268 & 31.887 & 1.966 & 23.775 &0.926& 0.416 & 0.263  &  0.814 & 0.744 &  1.000\\
            \hline
            ADE20k \rom{2} & \phantom{0}7,300& 983.4\tnote{f} & 16.187 & 307.65 &20.8\phantom{0}&  293.44 &1.000& 0.673 & 0.413   & 0.413  &0.846 & 0.900\\
            \hline
            ADE20k \rom{3} &12,193& 4,500\tnote{g}  & 13.202 & 263.217 & 51\phantom{.00}& 238.8  & 0.375& 0.937 & 0.375   & 0.375 & 0.930 & 0.937\\
            \hline
            ADE20k \rom{4} &47,468& 4,500\tnote{g} & 93.658 & 523.673  & 116\phantom{.0}& 423.349  & 0.375& NA & 0.608  &  0.608 &0.660  &  0.608\\

    \end{tabular}
    \begin{tablenotes}
        \item[a] DL : DL-Learner
        \item[b] DL FIC (1) : DL-Learner fast instance check with runtime capped at execution time of ECII DF
        \item[c] DL FIC (2) : DL-Learner fast instance check with runtime capped at execution time of ECII KCT
        \item[d] ECII DF : ECII default parameters 
        \item[e] ECII KCT : ECII keep common types and other default parameters 
        \item[f] Runtimes for DL-Learner were capped at 600 seconds. 
        \item[g] Runtimes for DL-Learner were capped at 4,500 seconds. 
    \end{tablenotes}
\end{threeparttable}
\end{adjustbox}
\end{center}
    \caption{Runtime and accuracy comparison of DL-Learner and ECII. Some figures are averages as described in the text.}
    \label{table:summary}
\end{table*}

We will now briefly discuss each of the scenarios in turn, before we summarize; Table \ref{table:summary} provides an overview of the results. Runtimes were averaged over 3 runs. Accuracy of a solution $S$ was assessed as
\begin{align*}
\alpha_3(S) &= \frac{|P_S| + |N_S|}{|P\cup N|},\text{ where}\\
P_S &= \{a\in P\mid \mathcal{O}\cup\mathcal{A}\models S(a)\} \text{ and}\\
N_S &=\{b\in N\mid \mathcal{O}\cup\mathcal{A}\not\models S(b)\},
\end{align*}
using a full reasoner. DL-Learner fast instance check (DL FIC) is compared with the $\alpha_2$ accuracy score (see above) of ECII system with default (ECII DF) and keeping common types (ECII KCT). In Table \ref{table:summary}, the accuracy value for DL-Learner is always the one with the best solution. The $\alpha_3$ accuracy score for DL-Learner is the score of the result returned by DL-Learner with the highest such score. The $\alpha_2$ score for ECII is the average score over three runs, where for each run the solution with the best $\alpha_2$ score is used for the average. The $\alpha_3$ accuracy score for ECII has been computed by taking all (or if more than five, a random section of five) of the results returned by ECII which score maximally in terms of $\alpha_2$. For each of these, the $\alpha_3$ score has been calculated using a full reasoner (in fact, Pellet), and averaging over these results.

The Yinyang family relationship problem is about creating descriptions for family relationship types from instance data. 
This dataset includes a small ontology with 157 logical axioms. 
We notice that ECII with default parametes performs worse than DL-Learner in this task both in terms of runtime and in terms of $\alpha_3$ accuracy, but ECII KCT obtains same $\alpha_2$ score as DL FIC when keeping the common types. 

The Michalski's train dataset 
ontology has 273 axioms. 
On this task, ECII is significantly outperformed by DL-Learner in terms of runtime, while both find perfect solutions. The reason that DL-Learner is quicker lies in the fact that it quickly comes up with solution by making a good choice in the refinement operator steps, thus leading to a quick termination. This is not necessarily an indication that DL-Learner runtime will always be significantly quicker for problems of this input size -- see e.g. the next paragraph.

The Forte family dataset is also of small size, the ontology has 341 axioms. The problem defined for this was to describe the uncle \cite{richards1995automated,LehmannH10} relationship. 
Two problems (large and small) were taken from this dataset by varying the numbers of positive and negative individuals, and the score is averaged. 
 ECII outperforms DL-Learner in terms of runtime but dips in accuracy. For approximation accuracy ECII ($\alpha_2=1.00$) outperforms DL-Learner fast instance check ($\alpha_2=0.87$) version by setting $keepCommonTypes$ to $true$. 

The Poker dataset has 1,368 axioms and thus a slightly larger ontology. \cite{LehmannH10} defined two problems using this dataset namely learning the definition of a pair and of a straight. With the same (1.0) accuracy, 
DL-Learner is significantly quicker, presumably for similar reasons as for the Trains dataset.

The Moral Reasoner dataset has 4,666 axioms and is of medium size as part of our set of scenarios. 
In \cite{LehmannH10}, several versions of this scenario were explored, and we chose the one using all instances and which did not modify the original set of atomic classes which can be used to construct solutions.
For this problem one of the solutions with accuracy 1.0 found by DL-Learner consists of the single atomic class ``guilty,'' while ECII average accuracy is only 0.785. One of the solutions found by ECII, with accuracy 0.91, is $ \neg \text{plan}\_\text{known} \sqcup  \neg \text{careful} \sqcup \text{guilty}$. DL-Learner 
was significantly quicker.
We see that DL-Learner outperforms ECII on this task, and the runtime performance of DL-Learner can be explained as in the Trains example. Regarding accuracy, ECII did not come up with ``guilty'' as candidate solution, presumably because its $\alpha_2$ score is not high enough. Indeed, the ontology for this dataset contains a significant numer of class disjunctions, which causes difficulties for our approximate reasoning procedure. This seems to explain both why ECII does not produce ``guilty'' as a solution and also why it would construct a complicated solution using disjunctions which scores well with respect to $\alpha_2$ but not well with respect to $\alpha_3$-accuracy.


The ADE20k dataset contains over 20,000 images of scenes which are preclassified in terms of scene type. Each image may have several annotations of objects which have been provided by humans. 
For our evaluation we used the dataset from \cite{SarkerXDRH17} where the ADE20k dataset, with annotations, was cast into an OWL ontology and aligned with the Standard Upper Merged Ontology SUMO\footnote{\url{http://www.adampease.org/OP/}}. In \cite{SarkerXDRH17}, sets of positive and negative exmples were selected from the images, and DL-Learner was used to generate corresponding class expressions. Only a handful of such experiments was reported in \cite{SarkerXDRH17}: As we see below and in Table \ref{table:summary}, the scenario yields to prohibitively long runtimes for DL-Learner, which makes a thorough investigation along the lines of \cite{SarkerXDRH17} impossible.


In order to compare performances, we used this scenario to create four problem classes with varying input sizes, as listed in Table \ref{table:summary}. 

For the smallest problem class, ontology size 4,714 axioms, 
ECII took on average 1.966 seconds to terminate and come up with solutions having average accuracy of 0.416. DL-Learner took 300 seconds to produce solutions with average accuracy of 0.926. As DL-Learner did not find solutions with accuracy 1.0 within a reasonable time span we capped execution at 
10 minutes or 600 seconds. DL-Learner then took on average 577.3 seconds and the average accuracy obtained within this time was 0.926 which is the same as of 300 seconds. 
For one representative problem, a solution
found by ECII 
had an accuracy of 1.0. 
The highest accuracy obtained by DL-Learner within this time limit on that same problem was 0.88.
We notice that accuracy of ECII lags behind that of DL-Learner on average, but with this input size we already see significant runtime improvements, and sometimes even higher accuracy. ECII outperforms DL FIC interms of $\alpha_2$ score.  

For the medium size setting -- ontology with 7,300 axioms, 
we used exactly the settings reported in \cite{SarkerXDRH17}. 
DL-Learner was able to come up with solutions with accuracy 1.0, but the time required to produce these solutions was 983 seconds average. The runtime for ECII was 20.8 seconds on average, and the average accuracy was 0.673. Accuracy of ECII on these is less than DL-Learner but runtime is very significantly improved. Approximate accuracy is better for ECII compared to DL FIC.

Our largest test ontologies were created using attributes from all images, from the types with names starting with  the letter ``R'' from the ADE20k dataset. In this case we created 2 different ontologies, in one version from the validation data -- ontology size 12,193 axioms -- and in the other version from the training data -- ontology size 47,468 axioms.

For the validation data ontology,
 DL-Learner was not able to find a solution with accuracy 1.0, and we terminated DL-Learner after 4,500 seconds (i.e., 75 minutes). DL-Learner produced a solution with accuracy of 0.375. ECII took 51 seconds to run and produced soultions with an average accuracy of 0.937. In fact, it turns out that DL-Learner did produce $\exists \text{imageContains}.\top$ as solution, which essentially means that it did not manage to take even a few refinement steps. One solution found by ECII 
had an accuracy of 0.90. ECII system also outperforms DL Fast instance check (DL FIC), in terms of $\alpha_2$ score. DL FIC was able to achive $\alpha_2$ score as 0.375 while ECII system achives significantly higher score.

For the training data ontology, 
DL-Learner was also terminated after 4,500 seconds. It procuded the same solution as before with accuracy of 0.375. ECII took 116 seconds producing an average $\alpha_2$-score of 0.66. Due to the size of the ontology, we were not able to run a reasoner to compute the $\alpha_3$ accuracy value for the solutions provided by ECII.

From the last two tasks, we see that ECII provides a very significant runtime improvement over DL-Learner, and is in fact able to produce approximate solutions in cases where DL-Learner can only return a trivial guess such as $\top$.



\begin{figure}[t]
    \centering
    \includegraphics[width=.5\textwidth]{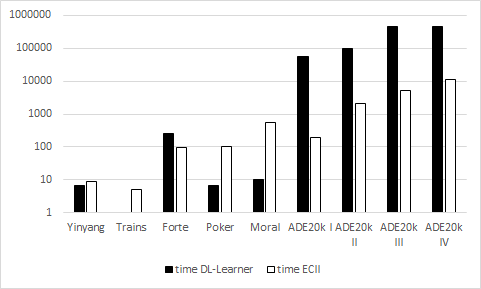}
    \caption{Runtime comparison between DL-Learner and ECII.  The vertical scale is logarithmic in hundredths of seconds, and note that DL-Learner runtime has been capped at 4,500 seconds for ADE20k III and IV. For ADE20k I it was capped at each run at 600 seconds.}
    \label{fig:runtime}
\end{figure}

Let us summarize the results using some charts. Figure~\ref{fig:runtime} displays a runtime comparison over all experiments; the experiments are sorted from left to right in increasing input size. The DL-Learner curve has much higher variance, which presumably is because runtime is capped whenever a 1.0 accuracy solution is found, while runtime is significantly higher for comparable sizes if this is not the case. ECII has an algorithm which is quicker in those test cases where DL-Learner did not find a 1.0 accuracy solution, i.e., for Forte and all ADE20k experiments, and is several orders of magnitudes quicker for the large input ontologies.

\begin{figure}[t]
    \centering
    \includegraphics[width=.5\textwidth]{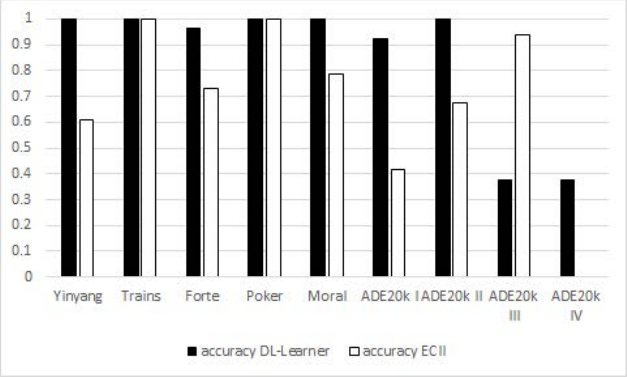}
    \caption{Accuracy ($\alpha_3$) comparison between DL-Learner and ECII. For ADE20k IV it was not possible to compute an accuracy score  within 3 hours for ECII as the input ontology was too large.}
    \label{fig:accuracy}
\end{figure}

\begin{figure}[t]
    \centering
    \includegraphics[width=.5\textwidth]{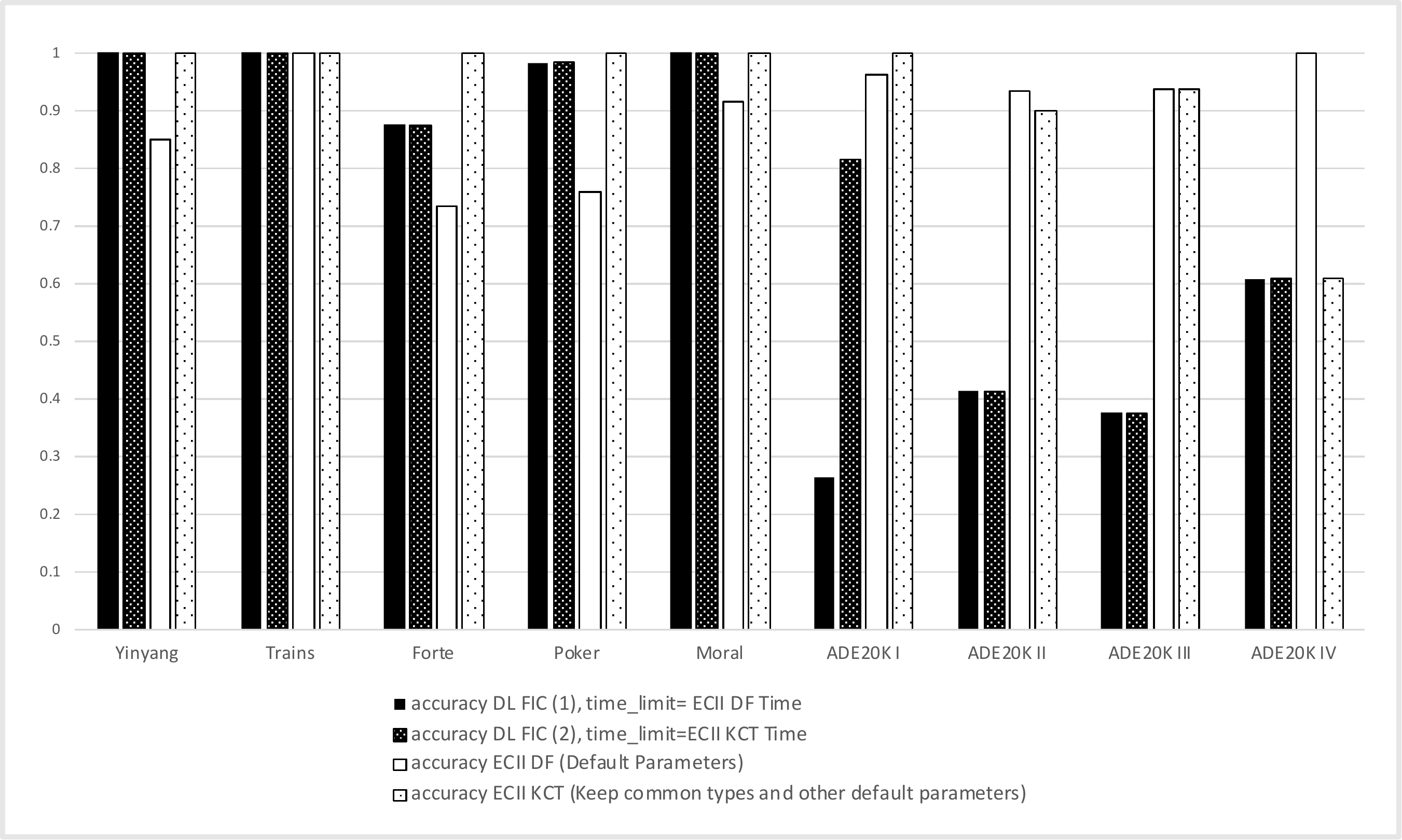}
    \caption{Approximation accuracy ($\alpha_2$) comparison between DL-Learner fast instance check and ECII. }
    \label{fig:accuracy-approximate}
\end{figure}

Figure \ref{fig:accuracy} visualizes the accuracy ($\alpha_3$) comparison results. While ECII achieves 1.0 accuracy, as does DL-Learner, in some cases (Train, Poker), it usually achieves only a lower accuracy for all cases where DL-Learner is able to produce non-trivial solutions. When DL-Learner produces only a trivial result, as in ADE20k III, ECII in fact is able to do better than DL-Learner. It is reasonable to assume that this would indeed happen in many cases where DL-Learner is simply not able to process a large input size.

Figure \ref{fig:accuracy-approximate} shows the approximate accuracy ($\alpha_2$) comparison results. We can see that ECII always outperforms DL FIC system. 



\section{Conclusions and Future Work}\label{sec:conc}

When we initially set out to develop the ECII algorithm and system, our goal was to provide an alternative to DL-Learner which trades some accuracy for speed. We anticipated that our approach would only be slightly less accurate but with  one or two orders of magnitude in runtime improvements.

In a sense, our experiments show that we were not bold enough in our assumptions regarding runtime improvmements for large input ontologies, while for smaller ones there aren't any, unless in the cases where DL-Learner cannot provide an accuracy 1.0 solution. At the same time, our experiments also show that we were too optimistic regarding accuracy results for the smaller ontologies, while at the same time we see much better results in the cases where DL-Learner has to resort to trivial solution attempts. 

So overall, based on the evaluation, ECII indeed seems to provide a reasonable alternative to DL-Learner in some cases, and in fact provides reasonable solutions even in cases where DL-Learner is unable to do so. From this perspective, we have achieved what we set out to do.

Our analysis of course also raises rather obvious points for further investigations and improvements. Further experiments in fact should shed further light on the strengths and weaknesses of ECII, and this needs to be explored. We have not varied the ECII default parameters, but we conjecture that we can move the time-accuracy trade-off in both directions by changing them. We have also not yet made full use of the theoretical results for OWL EL, but only looked at the obvious evaluation datasets for a fair comparison. 

As discussed in the evaluation section, it also turns out that the solutions with the highest $\alpha_2$-score are not always the best ones with respect to the correct $\alpha_3$-accuracy. Of course, calculating the $\alpha_3$-accuracy requires a full reasoner, and thus significant time, but by doing a few such checks one may be able to improve accuracy at the expense of some of the runtime gains. As a further alternative, a post-processing step could be added to the ECII algorithm which takes a somewhat larger number of the solution candidates which perform high with respect to $\alpha_2$-score, and returns only those among them which also score high on $\alpha_3$-accuracy.



\urlstyle{same}
\bibliographystyle{aaai} 
\bibliography{www12}
\end{document}